\def\year{2021}\relax
\documentclass[letterpaper]{article} 
\usepackage[switch]{lineno}  
\usepackage{aaai21}  
\usepackage{times}  
\usepackage{helvet} 
\usepackage{courier}  
\usepackage[hyphens]{url}  
\usepackage{graphicx} 
\usepackage{graphicx}  
\usepackage{natbib}  
\usepackage{caption} 
\usepackage{amsmath, amsfonts, amssymb, amsthm, xspace}
\usepackage{booktabs} 

\usepackage{subfigure, multirow, tabularx} 
\usepackage{booktabs} 
\usepackage{enumitem}
\usepackage{epstopdf}
\usepackage{graphicx}
\usepackage[noend,ruled,linesnumbered]{algorithm2e} 
\usepackage{microtype}
\usepackage{url}
\usepackage{lipsum}
\usepackage{makecell}

\title{Taxonomy Completion via Triplet Matching Network}
\author {
        Jieyu Zhang$^{\star}$,\textsuperscript{\rm 1}
        Xiangchen Song,\textsuperscript{\rm 2}
        Ying Zeng,\textsuperscript{\rm 3}
        Jiaze Chen,\textsuperscript{\rm 3}
        Jiaming Shen,\textsuperscript{\rm 2}
        Yuning Mao,\textsuperscript{\rm 2}
        Lei Li \textsuperscript{\rm 3} \\
}

\affiliations {
    \textsuperscript{\rm 1} Paul G. Allen School of Computer Science \& Engineering, University of Washington, WA, USA \\
    \textsuperscript{\rm 2} Department of Computer Science, University of Illinois Urbana-Champaign, IL, USA \\
    \textsuperscript{\rm 3} ByteDance AI Lab \\
    jieyuz2@cs.washington.edu, \{xs22, js2, yuningm2\}@illinois.edu, \{zengying.ss, chenjiaze, lileilab\}@bytedance.com
}

\newcommand{\ie}{\emph{i.e.}\xspace} 
\newcommand{\eg}{\emph{e.g.}\xspace} 
\newcommand{\mquote}[1]{{``\emph{#1}''}}

\newcommand{\naive}{na\"{\i}ve\xspace} 

\newtheorem{thm:def}{Definition}
\newtheorem{thm:eg}{Example}
\newtheorem{thm:lem}{Lemma}
\newtheorem{thm:obs}{Observation}
\newtheorem{thm:req}{Requirement}
\newtheorem{thm:prop}{Proposition}
\newtheorem{thm:principle}{Principle}
\newtheorem{thm:thm}{Theorem}
\newtheorem{thm:corollary}{Corollary}
\newtheorem{thm:property}{Property}

\newcommand{\pair}[1]{\langle #1 \rangle}			

\def \C {\mathcal{C}}
\def \D {\mathcal{D}}
\def \E {\mathcal{E}}

\def \N {\mathcal{N}}

\def \T {\mathcal{T}}

\newcommand{\TaxoExpan}{\mbox{\sf TMN}\xspace}
\newcommand{\TaxoExpanBf}{\mbox{\sf \textbf{TMN}}\xspace}

\begin{document}
\maketitle

{
\renewcommand{\thefootnote}{\fnsymbol{footnote}}
\footnotetext[1]{This work is done while interning at ByteDance AI Lab.}
}
\begin{abstract}
Automatically constructing taxonomy finds many applications in e-commerce and web search. One critical challenge is as data and business scope grow in real applications, new concepts are emerging and needed to be added to the existing taxonomy. Previous approaches focus on the taxonomy expansion, i.e. finding an appropriate hypernym concept from the taxonomy for a new query concept. In this paper, we formulate a new task, “taxonomy completion”, by discovering both the hypernym and hyponym concepts for a query. We propose \textbf{T}riplet \textbf{M}atching \textbf{N}etwork (\TaxoExpan\footnote{The code is released at \url{https://github.com/JieyuZ2/TMN}}), to find the appropriate $\langle$hypernym, hyponym$\rangle$ pairs for a given query concept. \TaxoExpan consists of one primal scorer and multiple auxiliary scorers. These auxiliary scorers capture various fine-grained signals (e.g., query to hypernym or query to hyponym semantics), and the primal scorer makes a holistic prediction on $\langle$query, hypernym, hyponym$\rangle$ triplet based on the internal feature representations of all auxiliary scorers. Also, an innovative channel-wise gating mechanism that retains task-specific information in concept representations is introduced to further boost model performance. Experiments on four real-world large-scale datasets show that \TaxoExpan achieves the best performance on both taxonomy completion task and the previous taxonomy expansion task, outperforming existing methods.
\end{abstract}

\section{Introduction}\label{sec:intro}

Taxonomies, formulated as directed acyclic graphs or trees, have been widely used to organize knowledge in various domains, such as news domain \cite{wikidata,mao2019hierarchical}, scientific domain \cite{Lipscomb2000MedicalSH, Sinha2015AnOO, shen2018entity} and online commerce \cite{txtract,mao2020octet}. 
Equipped with these curated taxonomies, researchers are able to boost the performance of numerous downstream applications such as query understanding~\cite{Hua2017UnderstandST, yang2020taxogan}, content browsing~\cite{Yang2012ConstructingTT}, personalized recommendation~\cite{Zhang2014TaxonomyDF, Huang2019TaxonomyAwareMR}, and web search~\cite{Wu2012ProbaseAP, Liu2019AUC}.  

\begin{figure}[!t]
  \centering
  \centerline{\includegraphics[scale=1.0, width=0.5\textwidth]{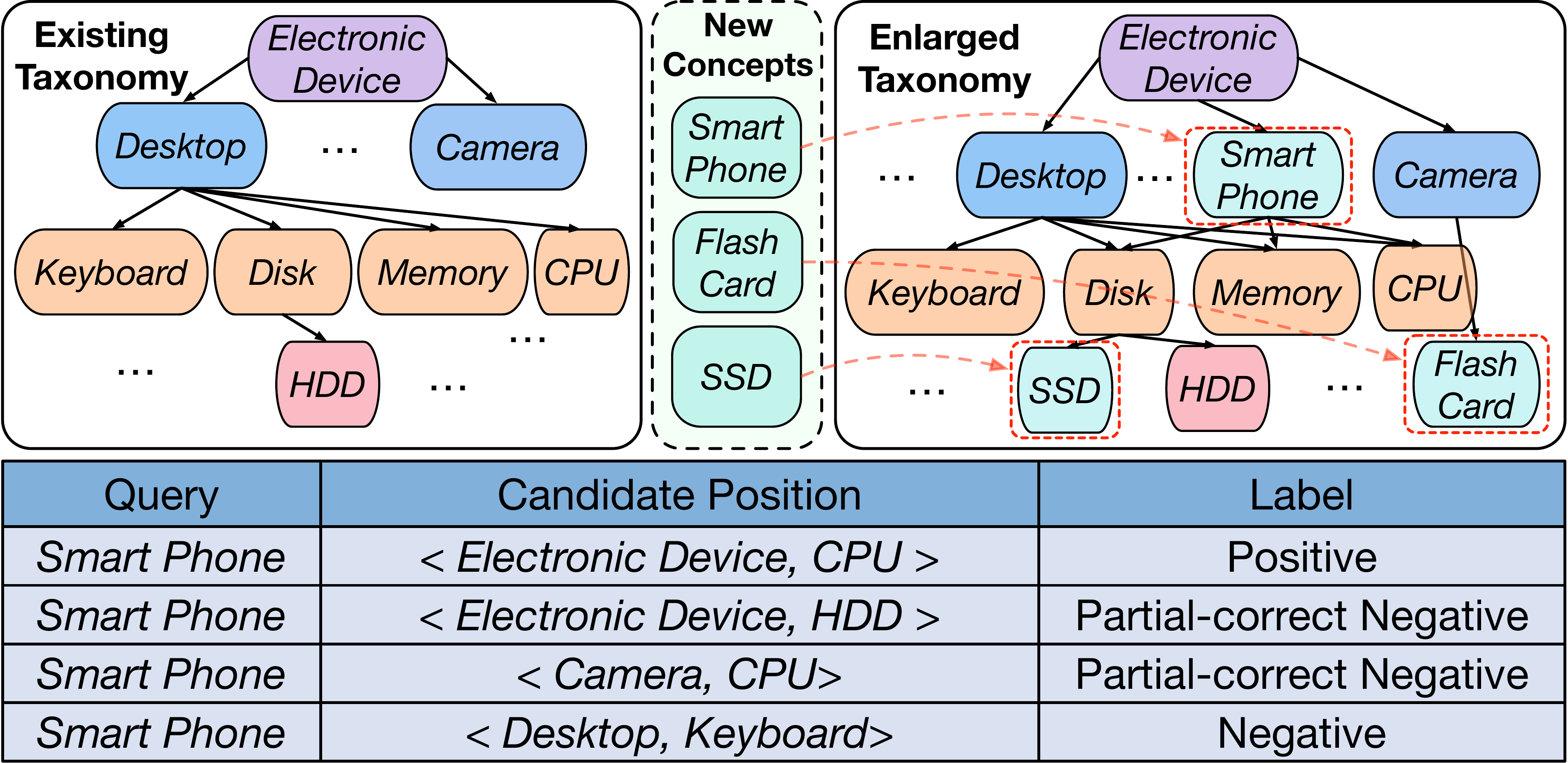}}
  \caption{An example of completing one \mquote{Electronic Device} taxonomy. The table illustrates different types of candidate positions for a given query \mquote{Smart Phone}.}
  \label{fig:example}
\end{figure}

As human knowledge is constantly growing and new concepts emerge everyday, it is needed to dynamically complete an existing taxonomy. Figure~\ref{fig:example} shows an illustrative example where a taxonomy of \mquote{Electronic Device} is completed to include new devices (\eg, \mquote{Smart Phone}) and hardware (\eg, \mquote{SSD}). Most existing taxonomies are curated by domain experts. However, such manual curations are labor-intensive, time-consuming and rarely-complete, and therefore infeasible to handle the influx of new contents in online streaming setting. 
To this end, many recent studies \cite{taxoexpan, manzoor2020expanding, steam} investigate the problem of \textit{taxonomy expansion} which aims to automatically expand an existing taxonomy. 
Specifically, given a query concept, these methods first rank each concept in the existing taxonomy based on how likely it is the hypernym of the query concept measured by an \textit{one-to-one} matching score between the two concepts. 
Then, the query concept is added into the existing taxonomy as the hyponym of the top-ranked concepts. 
Notice that such a formulation is built upon one strong assumption: all new concepts can only be added into existing taxonomy as hyponyms (i.e., leaf nodes\footnote{\small Nodes with zero out-degree in a directed acyclic graph.}).
However, we argue that such a ``hyponym-only'' assumption is inappropriate in real applications. For example, in Fig \ref{fig:example}, the term \mquote{Smart Phone} is invented much later than term \mquote{CPU}, which means that when \mquote{Smart Phone} emerges, \mquote{CPU} already exists in taxonomy. In this case, it is inappropriate to add \mquote{Smart Phone} into taxonomy as leaf node because \mquote{CPU} is a hyponym of \mquote{Smart Phone}. 

In this paper, instead, we define and investigate a new \textit{taxonomy completion} task \textit{without} the strong ``hyponym-only'' assumption. Formally, given an existing taxonomy and a set of new concepts, we aim to automatically complete the taxonomy to incorporate these new concepts by discovering the most likely $\langle$hypernym, hyponym$\rangle$ pairs of each new concept. For instance, in Fig \ref{fig:example}, one of the most likely candidate pairs for \mquote{Smart Phone} is $\langle$\mquote{Electronic Device}, \mquote{CPU}$\rangle$. 
This formulation leads to a novel \textit{one-to-pair} matching problem different from the previous \textit{one-to-one} setting in taxonomy expansion task that only seeks for a new concept's most likely hypernyms while ignores its possible hyponyms. 
Note that the hypernym/hyponym concept within the candidate $\langle$hypernym, hyponym$\rangle$ pair could be a “pseudo concept" in case there is no appropriate one for a given query concept. We can easily see that the taxonomy expansion task is a special case of taxonomy completion when the hyponym concepts are always ``pseudo concept''. 

Tackling the new taxonomy completion task is challenging because the induced one-to-pair matching problem results in the existence of a special type of negative candidate $\langle$hypernym, hyponym$\rangle$ pairs we called \textit{partially-correct negative candidates}. Before introducing partially-correct negative candidates, we first clarify that for a given query concept $n_q$, a candidate pair of existing concepts $\langle n_p, n_c\rangle$ is \textit{positive} if $n_p$($n_c$) is the true hypernym (hyponym) of $n_q$ and \textit{negtiave} otherwise. Then, a candidate pair $\langle n_p, n_c\rangle$ is partially-correct negative if either $n_p$ is true hypernym 
but $n_c$ is not true hyponym or vice versa. 
We illustrate the different types of candidate pairs in the table of Fig \ref{fig:example}.
Due to the high correlation of positive and partially-correct negative candidates, the model might struggle to distinguish one from another.

To solve the aforementioned challenge, we propose a novel \textbf{T}riplet \textbf{M}atching \textbf{N}etwork (\TaxoExpan), which learns a scoring function to output the matching score of a $\langle$query, hypernym, hyponym$\rangle$ triplet and leverages \textit{auxiliary signals} to help distinguish positive pairs from partially-correct negative ones. Specifically, auxiliary signals are binary signals indicating whether one component within the pair is positive or not, in contrast to binary \textit{primal signals} that reveal holistically whether a candidate position is positive or not. To make best use of the auxiliary signals to handle the existence of partially-correct negative, \TaxoExpan consists of multiple \textit{auxiliary scorers} that learn different auxiliary signals via corresponding auxiliary loss and one \textit{primal scorer} that aggregates internal feature representations of auxiliary scorers to output the final matching score. The auxiliary and primal scorers are jointly trained in an auxiliary learning framework. In this way, we encourage the model to learn meaningful internal feature representations for the primal scorer to discriminate between positive, negative, and partially-correct negative candidates. In addition, we propose an innovative technique called \textit{channel-wise gating mechanism} to regulate the representations of concepts. It produces a channel-wise gating vector based on the $\langle$query, hypernym, hyponym$\rangle$ triplet, and then modifies the embeddings using this channel-wise gating vector to reduce the effect of irrelevant information stored in embeddings while retain the most task-specific information when calculating matching scores. 

In the experiments, we benchmark the taxonomy completion task on four real-world taxonomies from different domains using modified version of multiple one-to-one matching models and state-of-the-art taxonomy expansion methods. Our experimental results show that \TaxoExpan outperforms the baselines by a large margin on both taxonomy completion task and taxonomy expansion task. Finally, ablation study demonstrates the effectiveness of each component of \TaxoExpan, and efficiency analysis shows the efficiency of \TaxoExpan at inference stage.

\noindent \textbf{Contributions.}
To summarize, our major contributions include:
(1) a more realistic task called taxonomy completion which simultaneously finds hypernym and hyponym concepts of new concepts; 
(2) a novel and effective Triple Matching Network (\TaxoExpan) to solve the one-to-pair matching problem induced from the taxonomy completion task by leveraging auxiliary signals and an innovative channel-wise gating mechanism; and
(3) extensive experiments that verify both the effectiveness and efficiency of \TaxoExpan framework on four real-world large-scale taxonomies from different domains.

\section{Problem Formulation}\label{sec:problem}

The input of the taxonomy completion task includes two parts: (1) an existing taxonomy $\T^{0} = (\N^{0}, \E^{0})$ and (2) a set of new concepts $\C$. We assume each concept $n_i \in \N^{0} \cup \C$ has an initial embedding vector $\mathbf{x}_i \in \mathbb{R}^d$ as in previous studies \cite{Jurgens2016SemEval2016T1, Vedula2018EnrichingTW, Aly2019EveryCS}. The overall goal is to complete the existing taxonomy $\T^{0}$ into a larger one $\T = (\N^{0} \cup \C, \E^{\prime})$ by adding the new concept $n_q \in \C$ between any pair of existing concepts $\langle n_{p}, n_{c}\rangle$ to form two new taxonomic relations $\langle n_{p}, n_q \rangle$ and $\langle n_q, n_{c} \rangle$. Following previous studies \cite{taxoexpan, manzoor2020expanding}, we assume the new concepts in $\C$ are independent to each other and thus reduce the more generic taxonomy completion task into $|C|$ independent simplified problems. Note that we use the term ``hypernym'' and ``parent'', ``hyponym'' and ``child'' interchangeably throughout the paper.

\noindent \textbf{Taxonomy.} Follow \cite{taxoexpan}, we define a taxonomy $\T = (\N, \E)$ as a directed acyclic graph where each node $n \in \N$ represents a concept (\ie, a word or a phrase) and each directed edge $\langle n_{p}, n_{c} \rangle \in \E$ indicates a relation expressing that concept $n_{p}$ is the most specific concept that is more general than concept $n_{c}$. Here, the relation types of edges are implicitly defined by existing taxonomy.

\noindent \textbf{Candidate Position.} A valid candidate position is a pair of concepts $\langle n_p, n_c\rangle$ where $n_c$ is one of the descendants of $n_p$ in the existing taxonomy. This definition reduces the search space of candidate positions. Note that $n_p$ or $n_c$ could be a ``pseudo concept'' acting as a placeholder.

\noindent \textbf{Positive Position.} For a query concept $n_q$, positive position $\langle n_p, n_c\rangle$ is a candidate position wherein $n_p$ and $n_c$ is the true parent and child of $n_q$, respectively.

\noindent \textbf{Negative Position.} For a query concept $n_q$, negative position $\langle n_p, n_c\rangle$ is a candidate position wherein $n_p$ or $n_c$ is not the true parent or child of $n_q$, respectively.

\noindent \textbf{Partially-correct Negative Position.} For a query concept $n_q$, partially-correct negative position $\langle n_p, n_c\rangle$ is a negative position but $n_p$ or $n_c$ is the true parent or child of $n_q$.
\section{The \TaxoExpan Framework}\label{sec:method}

\begin{figure*}[!t]
  \centering
  \centerline{\includegraphics[width=0.95\textwidth]{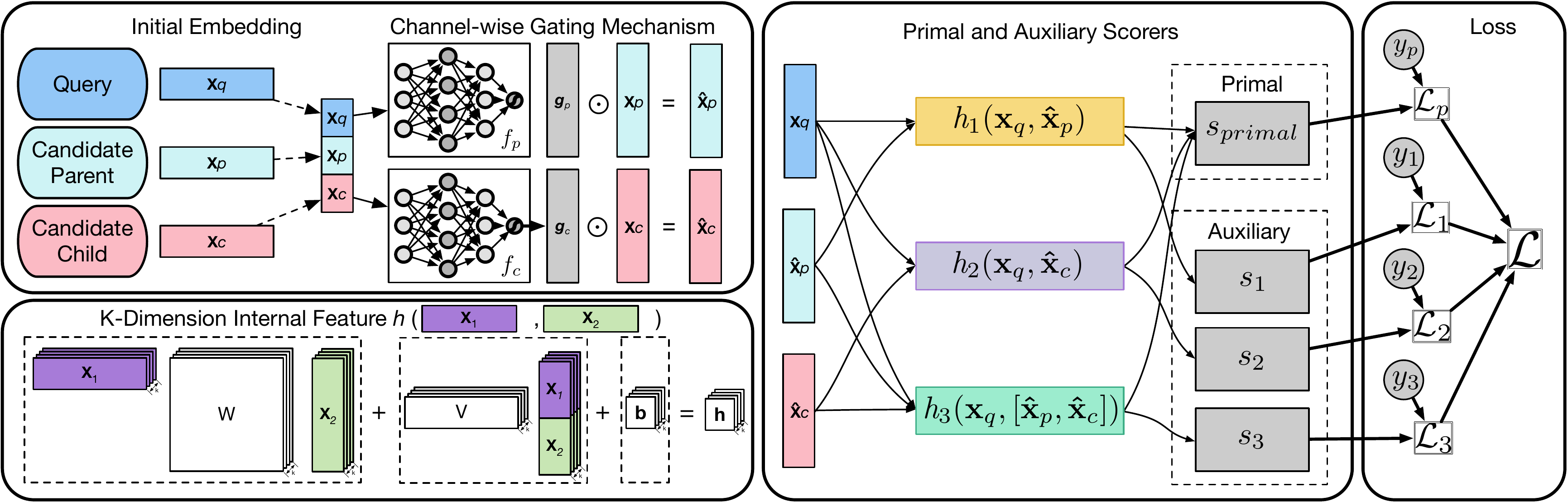}}
  \caption{Overview of \TaxoExpanBf framework.}
  \label{fig:overview}
\end{figure*}

In this section, we first introduce our one-to-pair matching model which leverages auxiliary signals to augment primal matching task.
Then, we present a novel channel-wise gating mechanism designed for regulating concept embedding to boost the model performance.
Finally, we discuss how to generate self-supervision data from the existing taxonomy and use them to train the \TaxoExpan model. The overall model architecture is presented in Fig \ref{fig:overview}.

 \subsection{Modeling One-To-Pair Matching}\label{subsec:decompose}

In this work, we seek to learn a model $s: \N \times (\N \times \N) \rightarrow \mathbb{R}$ with parameter $\Theta$ that can measure the relatedness of a query concept $n_q$ and a candidate position, \ie, a pair of concepts $t=\langle n_{p}, n_{c}\rangle$, in existing taxonomy $\T^{0}$. A straightforward instantiation of $s$ is as follows:
\begin{equation}\label{naive}
    \small
s(n_q, n_{p}, n_{c}) = f(\mathbf{x}_q, [\mathbf{x}_p, \mathbf{x}_c]) = f(\mathbf{x}_q, \mathbf{x}_t)
\end{equation}
Where $f$ is a parameterized scoring function of choice that outputs the relatedness score of $n_q$ and $\langle n_{p}, n_{c}\rangle$, and $[\cdot]$ represents the concatenation operation. This formulation simply degenerates one-to-pair matching into one-to-one matching by using concatenation of $\mathbf{x}_p$ and $\mathbf{x}_c$ as representation of candidate position. Here, we choose the neural tensor network \cite{socher2013reasoning} as our base model:
\begin{equation}\label{ntn}
\small
s(n_q, n_{p}, n_{c}) = \mathbf{u}^{T}\sigma(h(\mathbf{x}_q, \mathbf{x}_t)
)
\end{equation}
\begin{equation}\label{h}
\small
h(\mathbf{x}_q, \mathbf{x}_t) = \mathbf{x}_q \mathbf{W}^{[1:k]} \mathbf{x}_t + \mathbf{V} 
\left[\begin{aligned}
    \mathbf{x}_q\\ 
    \mathbf{x}_t\\  
\end{aligned}\right] + \mathbf{b}
\end{equation}
Where $\sigma=\tanh(\cdot)$ is a standard nonlinearity applied element-wise, $\mathbf{W}^{[1:k]} \in \mathbb{R}^{d\times 2d \times k}$ is a tensor and the bilinear tensor product $\mathbf{x}_q \mathbf{W}^{[1:k]} \mathbf{x}_t$ results in a vector $\mathbf{r} \in \mathbb{R}^k$, where each entry is computed by one slice $i = 1, \dots, k$ of the tensor: $\mathbf{h}_i = \mathbf{x}_q \mathbf{W}^{i} \mathbf{x}_t$. The other parameters  are the standard form of a neural network: $\mathbf{V} \in \mathbb{R}^{k \times 2d}$ and $\mathbf{u}\in \mathbb{R}^{k}$, $\mathbf{b}\in \mathbb{R}^{k}$. We call the vector $\mathbf{h}$ output by $h(\cdot, \cdot)$ the internal feature representation of neural tensor network.

However, such a \naive instantiation only measures the coarse-grained relatedness of query $n_q$ and the whole candidate pair $\langle n_p, n_c\rangle$ but fails to capture the fine-grained relatedness of $\langle n_q, n_p\rangle$ and $\langle n_q, n_c\rangle$,  preventing the model from learning to clearly distinguish positive candidates from partially-correct negatives candidates. 

To address the limitation of the naive approach, we propose a novel expressive \textbf{Triplet Matching Network (\TaxoExpan)}. Specifically, we develop multiple auxiliary scorers to capture both coarse- and fine-grained relatedness in one-to-pair matching, and one primal scorer that inputs the internal feature representations of all auxiliary scorers and outputs final matching scores. For each auxiliary scorer, we adopt neural tensor network as in Eq. \ref{ntn} as instantiation due to its expressiveness, and the corresponding $k$-dimension internal feature representation $\mathbf{h}$ is as in Eq. \ref{h}. Assume we have $l$ auxiliary scorers, each with $k$-dimension internal feature representation $\mathbf{h}_j$ and $j = 1, \dots, l$. Then, the primal scorer is a single-layer projection with non-linear activation function:
\begin{equation}\label{primal}
\small
s_{primal}(n_q, n_{p}, n_{c}) = \mathbf{u}_{p}^{T}\sigma([\mathbf{h}_{1}, \dots, \mathbf{h}_{l}])
\end{equation}
Where, again, $\sigma=\tanh(\cdot)$ and $\mathbf{u}_p \in \mathbb{R}^{kl}$. Now we elaborate the three auxiliary scorers that capture both coarse- and fine-grained relatedness:
\begin{equation}\label{s1}
\small
s_{1}(n_q, n_{p}) = \mathbf{u}_{1}^{T}\sigma(\mathbf{h}_{1}(\mathbf{x}_q, \mathbf{x}_p))
\end{equation}
\begin{equation}\label{s2}
\small
s_{2}(n_q, n_{c}) = \mathbf{u}_{2}^{T}\sigma(\mathbf{h}_{2}(\mathbf{x}_q, \mathbf{x}_c))
\end{equation}
\begin{equation}\label{s3}
\small
s_{3}(n_q, n_p, n_{c}) = \mathbf{u}_{3}^{T}\sigma(\mathbf{h}_{3}(\mathbf{x}_q, [\mathbf{x}_p, \mathbf{x}_c])) 
\end{equation}
Where the auxiliary scorer $s_{1}$ and $s_{2}$ capture the fine-grained relatedness of $\langle n_q, n_p\rangle$ and $\langle n_q, n_c\rangle$ respectively, while $s_{3}$ is for coarse-grained relatedness between $n_q$ and $\langle n_p, n_c\rangle$.

Given above formulations, primal scorer $s_{primal}$ can be trained using primal signals indicating whether $\langle n_p, n_c\rangle$ is positive candidate of $n_q$ or not, and auxiliary scorers can be trained via corresponding auxiliary signals. Particularly, $s_1$ will be trained to learn whether $n_p$ is positive parent of $n_q$, $s_2$ is to learn whether $n_c$ is positive child of $n_q$, and $s_3$ captures coarse-grained relatedness between $n_q$ and $\langle n_p, n_c\rangle$ so its auxiliary signal is exactly the same as primal signal. Although $s_3$ and $s_{primal}$ share the same supervision signals and both aim to capture relatedness between $n_q$ and $\langle n_p, n_c\rangle$, $s_{primal}$ outputs matching score based on the internal feature representations of all auxiliary scorers including $s_3$, which enables $s_{primal}$ to rely on $s_1$ or $s_2$ when $s_3$ struggle to differentiate positive candidates from partially-correct candidates negatives.  

\subsection{Channel-wise Gating Mechanism}\label{subsec:gating}

As the nature of taxonomy, concepts under the same ancestor are semantically related to each other, which makes it challenging for model to learn the true taxonomic relations based on concept embeddings, especially in bottom-level of a taxonomy. For instance, in Fig~\ref{fig:example}, the model needs to learn that \mquote{Disk} is the true parent of \mquote{SSD} but \mquote{Memory} is not. However, \mquote{Disk} and \mquote{Memory} are siblings in taxonomy, which makes them highly-related, and therefore hard to distinguish based on their embeddings learned from a more general corpus. 

To mitigate this problem, instead of directly using initial embedding vectors of concepts, we propose a novel channel-wise gating mechanism to regulate the information stored in initial embedding vectors, reducing the negative effects of irrelevant or spurious information on learning taxonomic relations. Specially, to distinguish \mquote{Disk} and \mquote{Memory} that both belong to \mquote{Desktop}, we would like to filter out the shared information stored in their embeddings related to \mquote{Desktop} in order to push the model to focus on the remaining more specific information. Formally, we give the formulation of channel-wise gating mechanism as follows:
\begin{equation}\label{gp}
\small
\mathbf{g}_p = \theta(\mathbf{W}_{1}[\mathbf{x}_q, \mathbf{x}_p, \mathbf{x}_c]) 
\end{equation}
\begin{equation}\label{gatep}
\small
\hat{\mathbf{x}}_p = \mathbf{g}_p \odot \mathbf{x}_p
\end{equation}
\begin{equation}\label{gc}
\small
\mathbf{g}_c = \theta(\mathbf{W}_{2}[\mathbf{x}_q, \mathbf{x}_p, \mathbf{x}_c]) 
\end{equation}
\begin{equation}\label{eq:gatec}
\small
\hat{\mathbf{x}}_c = \mathbf{g}_c \odot \mathbf{x}_c
\end{equation}
Where $\theta(\cdot)$ is a sigmoid function and $\mathbf{W}_1,  \mathbf{W}_2 \in \mathbb{R}^{d \times 3d}$. $\odot$ is element-wise multiplication. $\mathbf{g}$ is the channel-wise gating vector dependent on embeddings of both query and candidate positions. We treat each dimension of concept embedding as a channel to store information, and the output value of sigmoid lies in the interval $[0, 1]$, which enables the channel-wise gating vector $\mathbf{g}$ to downscale irrelevant information in certain channels while retain task-specific ones. 

With the gated embedding $\hat{\mathbf{x}}_p $ and $\hat{\mathbf{x}}_c $ in hand, we now replace the initial embedding vectors in Eq. \ref{h} with it to facilitate \TaxoExpan. Notably, this simple channel-wise gating mechanism is ready to be plugged in any matching models.

\subsection{Jointly Learning Primal and Auxiliary Scorers}\label{subsec:learning}
In this section, we first introduce how to generate self-supervision data as well as primal and auxiliary signals from the existing taxonomy, and then propose to jointly learn the primal and auxiliary scorers.

\smallskip
\noindent \textbf{Self-supervision Generation.} 
Given one node $n_q$ in the existing taxonomy $\T^{0} = (\N^{0}, \E^{0})$ as \mquote{query}, we first construct a positive $\pair{n_p, n_c}$ pair by using one of its parents $n_p$ and one of its children $n_c$. Then, we sample $N$ negative pairs from valid candidates positions that are not positive positions. Notably, it is allowed that one end of the negative pair is true parent/child of $n_q$.
 Given a candidate pair $\pair{n_p, n_c}$, the primal signal, \ie $y$, indicates whether the whole pair is positive or not, while the auxiliary signals, \ie $y_p$ and $y_c$, represent whether $n_p$ ($n_c$) is the true parent (child) of query $n_q$ respectively regardless of the primal signal. The generated positive and negative pairs as well as corresponding primal and auxiliary signals collectively consist of one training instance ($\mathbf{X}$, $\mathbf{y}$). By repeating the above process for each node in $\T^{0}$, we obtain the full self-supervision dataset $\mathbb{D} = \{(\mathbf{X_1}, \mathbf{y_1}), \dots, (\mathbf{X_{|\N^{0}|}}, \mathbf{y_{|\N^{0}|}}) \}$.

\smallskip
\noindent \textbf{Learning Objective.}
We learn our model on $\mathbb{D}$ using the following objective:
\begin{equation}\label{eq:loss general}
\small
\mathcal{L}(\Theta) =\mathcal{L}_{p} + \lambda_1\mathcal{L}_{1} + \lambda_2\mathcal{L}_{2} + \lambda_3\mathcal{L}_{3} 
\end{equation}
where $\mathcal{L}_p$ represents the loss for primal scorer and $\mathcal{L}_1$, $\mathcal{L}_2$, $\mathcal{L}_3$ are auxiliary losses for auxiliary scorers $s_1$, $s_2$ and $s_3$ respectively. The hyperparameters $\lambda_1$, $\lambda_2$ and $\lambda_3$ are weights to adjust relative effects of each auxiliary loss. The above objective function is similar to multi-task learning at the first glance, but it is an auxiliary learning strategy that only cares the performance of primal task, \ie primal scorer in our case, and the auxiliary loss are meant to augment the learning of primal task.

Here, $\mathcal{L}_i$ is loss function of choice. We choose binary cross entropy loss for simplicity. Take the primal loss as an example, it is formulated as:
\begin{equation}\label{eq:loss}
\small
\mathcal{L}_{p} = -\frac{1}{|\mathbb{D}|} \sum_{(\mathbf{X}_i, y_i) \in \mathbb{D}}  y_i \log(s_{p}(\mathbf{X}_i)) + (1-y_i) \log(1-s_{p}(\mathbf{X}_i))  
\end{equation}
\section{Experiments}\label{sec:exp}
\begin{table}[t]
    \centering
   
    \scalebox{0.9}{
        \begin{tabular}{ccccc}
            \toprule
            \textbf{Dataset}    & $|\N|$ &  $|\E|$ & $|\D|$ \\
            \midrule
            \textbf{MAG-CS}      & 24,754     &  42,329   & 6  \\
            \textbf{MAG-Psychology}    & 23,187    &   30,041   &   6  \\
            \textbf{WordNet-Verb}   & 13,936      & 13,408      & 13  \\
            \textbf{WordNet-Noun}   & 83,073      & 76,812      & 20  \\
    	\bottomrule
        \end{tabular}
    }
    \caption{Dataset Statistics. $|\N|$ and $|\E|$ are the number of nodes and edges in the existing taxonomy. $|\D|$ indicates the taxonomy depth.}
    \label{tbl:dataset}
\end{table}

\begin{table*}[!t]
	\centering
	\scalebox{0.7}{
        \begin{tabular}{c|cccccccc}
        		\toprule
             	\multirow{2}{*}{\textbf{Method}} & \multicolumn{8}{c}{\textbf{MAG-CS}} \\
		\cmidrule{2-9}
		& MR & MRR & Recall@1 & Recall@5 & Recall@10  & Prec@1 & Prec@5 & Prec@10   \\
		\midrule
		Closest-Position &9466.670 & 0.093 & 0.012 & 0.034 & 0.051&  0.054&  0.029 &0.022\\
		Single Layer Model  & 1025.245 $\pm$ 22.827 & 0.153 $\pm$ 0.002  & 0.030 $\pm$ 0.001  & 0.074 $\pm$ 0.001  & 0.105 $\pm$ 0.002  & 0.130 $\pm$ 0.004  & 0.064 $\pm$ 0.001  & 0.046 $\pm$ 0.001\\
		Multiple Layer Model  & 1110.340 $\pm$ 64.987 & 0.156 $\pm$ 0.002 & \textbf{0.032} $\pm$ 0.001	& \textbf{0.080} $\pm$ 0.001	& 0.108 $\pm$ 0.002	& \textbf{0.140} $\pm$ 0.004	& \textbf{0.069} $\pm$ 0.000	& 0.047 $\pm$ 0.001\\
		Bilinear Model          &3373.772 $\pm$ 7.473 &0.026 $\pm$ 0.000&	0.000 $\pm$ 0.000&	0.003 $\pm$ 0.000&	0.006 $\pm$ 0.000&	0.001 $\pm$ 0.000&	0.002 $\pm$ 0.000&	0.003 $\pm$ 0.000\\
		Neural Tensor Network   &\textbf{769.830} $\pm$ 2.416&\textbf{0.171} $\pm$ 0.003& 0.023 $\pm$ 0.001&	0.073 $\pm$ 0.001&	\textbf{0.112} $\pm$ 0.002&	0.099 $\pm$ 0.004&	0.063 $\pm$ 0.001&	\textbf{0.049} $\pm$ 0.001	\\
TaxoExpan & 1523.904 $\pm$ 52.982   & 0.099 $\pm$ 0.002 & 0.004 $\pm$ 0.001  & 0.027 $\pm$ 0.002  & 0.049 $\pm$ 0.001  & 0.017 $\pm$ 0.003  & 0.023 $\pm$ 0.001  & 0.021 $\pm$ 0.000 \\
		ARBORIST  & 1142.335 $\pm$ 19.249 & 0.133 $\pm$ 0.004  & 0.008 $\pm$ 0.001  & 0.044 $\pm$ 0.003  & 0.075 $\pm$ 0.003  & 0.037 $\pm$ 0.004  & 0.038 $\pm$ 0.003  & 0.033 $\pm$ 0.001  \\
		\midrule
 		\TaxoExpan     &\textbf{639.126} $\pm$ 35.849&	\textbf{0.204} $\pm$ 0.005 &\textbf{0.036} $\pm$ 0.003&	\textbf{0.099} $\pm$ 0.006&	\textbf{0.139} $\pm$ 0.006&	\textbf{0.156} $\pm$ 0.011&	\textbf{0.086} $\pm$ 0.005&	\textbf{0.060} $\pm$ 0.003 \\
		\bottomrule
         \end{tabular}
 	}
 	\scalebox{0.7}{
        \begin{tabular}{c|cccccccc}
        		\toprule
             	\multirow{2}{*}{\textbf{Method}} & \multicolumn{8}{c}{\textbf{MAG-Psychology}} \\
		\cmidrule{2-9}
		& MR & MRR & Recall@1 & Recall@5 & Recall@10  & Prec@1 & Prec@5 & Prec@10   \\
		\midrule
		Closest-Position &5201.604 &0.168 &0.030 &0.072 &0.107 & 0.062 & 0.029 &0.022\\
		Single Layer Model  & 435.548 $\pm$ 4.057  & 0.350 $\pm$ 0.002  & 0.090 $\pm$ 0.003  & 0.209 $\pm$ 0.002  & 0.274 $\pm$ 0.003  & 0.183 $\pm$ 0.006  & 0.085 $\pm$ 0.001  & 0.056 $\pm$ 0.001  \\
		Multiple Layer Model  & \textbf{297.644} $\pm$ 8.097	& \textbf{0.413} $\pm$ 0.002& \textbf{0.110} $\pm$ 0.001	& \textbf{0.265} $\pm$ 0.004	& \textbf{0.334} $\pm$ 0.002	& \textbf{0.224} $\pm$ 0.003	& \textbf{0.108} $\pm$ 0.001	& \textbf{0.068} $\pm$ 0.000\\
		Bilinear Model        & 2113.024 $\pm$ 9.231	&	0.032 $\pm$ 0.000& 0.000 $\pm$ 0.000&	0.001 $\pm$ 0.000&	0.003 $\pm$ 0.000&	0.000 $\pm$ 0.000&	0.000 $\pm$ 0.000&	0.001 $\pm$ 0.000 \\
		Neural Tensor Network   &299.004 $\pm$ 6.294&	0.380 $\pm$ 0.004&0.066 $\pm$ 0.004&	0.207 $\pm$ 0.002&	0.291 $\pm$ 0.004&	0.134 $\pm$ 0.008&	0.084 $\pm$ 0.001	&0.059 $\pm$ 0.001\\
TaxoExpan & 728.725 $\pm$ 2.096  & 0.253 $\pm$ 0.001 & 0.015 $\pm$ 0.001  & 0.092 $\pm$ 0.001  & 0.163 $\pm$ 0.001  & 0.031 $\pm$ 0.001  & 0.038 $\pm$ 0.000  & 0.033 $\pm$ 0.000  \\
		ARBORIST &547.723 $\pm$ 20.165   & 0.344 $\pm$ 0.012 & 0.062 $\pm$ 0.009  & 0.185 $\pm$ 0.011  & 0.256 $\pm$ 0.013  & 0.126 $\pm$ 0.018  & 0.076 $\pm$ 0.004  & 0.052 $\pm$ 0.003 \\
		\midrule
 		\TaxoExpan     &\textbf{212.298} $\pm$ 3.051 &	\textbf{0.471} $\pm$ 0.001& \textbf{0.141} $\pm$ 0.001&	\textbf{0.305} $\pm$ 0.004&	\textbf{0.377} $\pm$ 0.002&	\textbf{0.287} $\pm$ 0.001&	\textbf{0.124} $\pm$ 0.001&	\textbf{0.077} $\pm$ 0.000\\
		\bottomrule
         \end{tabular}
    }
    \scalebox{0.7}{
        \begin{tabular}{c|cccccccc}
        		\toprule
             	\multirow{2}{*}{\textbf{Method}} & \multicolumn{8}{c}{\textbf{WordNet-Verb}} \\
		\cmidrule{2-9}
		& MR & MRR & Recall@1 & Recall@5 & Recall@10  & Prec@1 & Prec@5 & Prec@10   \\
		\midrule
		Closest-Position &34778.772&0.144&0.011&0.045&0.075&0.020&0.016&0.013\\
		Single Layer Model  & 2798.243 $\pm$ 61.384  & 0.140 $\pm$ 0.009 & 0.029 $\pm$ 0.005  & 0.065 $\pm$ 0.006  & 0.093 $\pm$ 0.008  & 0.044 $\pm$ 0.007  & 0.019 $\pm$ 0.002  & 0.014 $\pm$ 0.001\\
		Multiple Layer Model  & 2039.213 $\pm$ 240.577	& 0.227 $\pm$ 0.020	& 0.050 $\pm$ 0.006	& 0.120 $\pm$ 0.009	& 0.160 $\pm$ 0.015	& 0.075 $\pm$ 0.009	& 0.036 $\pm$ 0.003	& 0.024 $\pm$ 0.002\\
		Bilinear Model         & 1863.915 $\pm$ 5.685 &	0.175 $\pm$ 0.001&	0.012 $\pm$ 0.001&	0.054 $\pm$ 0.000&	0.096 $\pm$ 0.001&	0.017 $\pm$ 0.001	&0.016 $\pm$ 0.000	&0.015 $\pm$ 0.000\\
		Neural Tensor Network   &\textbf{1599.196} $\pm$ 18.409	&\textbf{0.255} $\pm$ 0.003&	\textbf{0.051} $\pm$ 0.002&	\textbf{0.125} $\pm$ 0.006&	\textbf{0.176} $\pm$ 0.005&	\textbf{0.076} $\pm$ 0.003&	\textbf{0.038} $\pm$ 0.002&	\textbf{0.027} $\pm$ 0.001\\
TaxoExpan & 1799.939 $\pm$ 4.511 & 0.227 $\pm$ 0.002  & 0.024 $\pm$ 0.001  & 0.095 $\pm$ 0.001  & 0.140 $\pm$ 0.002  & 0.036 $\pm$ 0.002  & 0.029 $\pm$ 0.000  & 0.021 $\pm$ 0.000  \\
			ARBORIST & 1637.025 $\pm$ 4.950  & 0.206 $\pm$ 0.011 & 0.016 $\pm$ 0.004  & 0.073 $\pm$ 0.011  & 0.116 $\pm$ 0.011  & 0.024 $\pm$ 0.006  & 0.022 $\pm$ 0.003  & 0.018 $\pm$ 0.002 \\
			\midrule
 		\TaxoExpan     & \textbf{1445.801} $\pm$ 27.209  & \textbf{0.304} $\pm$ 0.005 & \textbf{0.072} $\pm$ 0.003  & \textbf{0.163} $\pm$ 0.005  & \textbf{0.215} $\pm$ 0.001  & \textbf{0.108} $\pm$ 0.005  & \textbf{0.049} $\pm$ 0.002  & \textbf{0.032} $\pm$ 0.000 \\
		\bottomrule
         \end{tabular}
    }
    \scalebox{0.7}{
        \begin{tabular}{c|cccccccc}
        		\toprule
             	\multirow{2}{*}{\textbf{Method}} & \multicolumn{8}{c}{\textbf{WordNet-Noun}} \\
		\cmidrule{2-9}
		& MR & MRR & Recall@1 & Recall@5 & Recall@10  & Prec@1 & Prec@5 & Prec@10   \\
		\midrule
		Closest-Position &5601.033&0.136&0.017&0.044&0.074&0.025&0.013&0.011\\
		Single Layer Model  & 3260.415 $\pm$ 79.776 & 0.177 $\pm$ 0.010 & 0.025 $\pm$ 0.003  & 0.072 $\pm$ 0.006  & 0.103 $\pm$ 0.006  & 0.043 $\pm$ 0.005  & 0.025 $\pm$ 0.002  & 0.018 $\pm$ 0.001 \\
		Multiple Layer Model  & \textbf{2801.500} $\pm$ 143.579		& 0.175 $\pm$ 0.005& 0.029 $\pm$ 0.001	& 0.077 $\pm$ 0.002	& 0.106 $\pm$ 0.003	& 0.051 $\pm$ 0.002	& 0.027 $\pm$ 0.001	& 0.018 $\pm$ 0.000\\
		Bilinear Model         &3498.184 $\pm$ 3.586&	0.176 $\pm$ 0.001&	0.012 $\pm$ 0.000&	0.052 $\pm$ 0.001&	0.095 $\pm$ 0.001&	0.020 $\pm$ 0.000&	0.018 $\pm$ 0.000&	0.017 $\pm$ 0.000 \\
		Neural Tensor Network   &2808.900 $\pm$ 79.415&	0.215 $\pm$ 0.007& \textbf{0.034} $\pm$ 0.002&	\textbf{0.093} $\pm$ 0.003&	\textbf{0.133} $\pm$ 0.004&	\textbf{0.060} $\pm$ 0.004&	\textbf{0.033} $\pm$ 0.001&	\textbf{0.023} $\pm$ 0.001\\
		TaxoExpan & 3188.935 $\pm$ 17.461 & 0.209 $\pm$ 0.000  & 0.017 $\pm$ 0.000 & 0.074 $\pm$ 0.000 & 0.125 $\pm$ 0.001 & 0.030 $\pm$ 0.001 & 0.026 $\pm$ 0.000 & 0.022 $\pm$ 0.000 \\
	    ARBORIST & 2993.341 $\pm$ 114.749 & \textbf{0.217} $\pm$ 0.005  & 0.021 $\pm$ 0.001  & 0.073 $\pm$ 0.002  & 0.125 $\pm$ 0.002  & 0.036 $\pm$ 0.001  & 0.025 $\pm$ 0.001  & 0.022 $\pm$ 0.000  \\
	    \midrule
 		\TaxoExpan      &\textbf{1647.665} $\pm$ 15.370 & \textbf{0.270} $\pm$ 0.006&\textbf{0.039} $\pm$ 0.002&	\textbf{0.111} $\pm$ 0.006&	\textbf{0.167} $\pm$ 0.005	&\textbf{0.068} $\pm$ 0.004	&\textbf{0.039} $\pm$ 0.002&	\textbf{0.029} $\pm$ 0.001	\\
		\bottomrule
         \end{tabular}
    }
    \caption{Overall results on four datasets. We run all methods five times and report the averaged result with standard deviation. Note that smaller MR indicates better model performance. For all other metrics, larger values indicate better performance. We highlight \underline{the best two models} in terms of the average performance under each metric.}
    \label{tbl:overall_results}
\end{table*}

\subsection{Experimental Setup}
\smallskip
\noindent \textbf{Dataset.} We study the performance of \TaxoExpan on four large-scale real-world taxonomies.
\begin{itemize}[leftmargin=*]
\item \noindent \textbf{Microsoft Academic Graph (MAG).} We evaluate \TaxoExpan on the public Field-of-Study (FoS) Taxonomy in Microsoft Academic Graph (MAG)~\cite{Sinha2015AnOO}.  It contains over 660 thousand scientific concepts and more than 700 thousand taxonomic relations. Following \cite{taxoexpan}, we construct two datasets which we refer to as \textbf{MAG-Psychology} and \textbf{MAG-CS} based on the subgraph related to the ``Psychology'' and ``Computer Science'' domain, respectively. We compute a 250-dimension word word2vec embedding on a related paper abstracts corpus.
 \item \noindent \textbf{WordNet.}  Based on WordNet 3.0, we collect verbs and nouns along with the relations among them to form two datasets which we refer to as \textbf{WordNet-Verb} and \textbf{WordNet-Noun}, respectively. The reason for limiting our choice to only verbs and nouns is that only these parts of speech have fully-developed taxonomies in WordNet \cite{Jurgens2016SemEval2016T1}. We obtain the 300-dimension fasttext embeddings\footnote{\scriptsize We use the wiki-news-300d-1M-subword.vec.zip version on official website.} as initial feature vectors. 
 \end{itemize}
 
For each dataset, we randomly sample 1,000 nodes for validation and another 1,000 for test. Then we build the initial taxonomy using remaining nodes and associated edges. Notice that new edges will be added into initial taxonomy to avoid the taxonomy from breaking into multiple directed acyclic graphs.
Table~\ref{tbl:dataset} lists the statistics of these four datasets.

\smallskip
\noindent \textbf{Evaluation Metrics}. As our model returns a rank list of candidate positions for each query concept, we evaluate its performance using the following ranking-based metrics.
\begin{itemize}[leftmargin=*, itemsep=0mm]
\item \textbf{Mean Rank (MR)} measures the average rank position of a query concept's true positions among all candidates. For queries with multiple positive edges, we first calculate the rank position of each individual edge and then take the average of all rank positions.

\item \textbf{Mean Reciprocal Rank (MRR)} calculates the reciprocal rank of a query concept's true positions. We follow \cite{Ying2018GraphCN} and scale the original MRR by a factor 10 to amplify the performance gap between different methods. 

\item \textbf{Recall@}$k$ is the number of query concepts' true positions ranked in the top $k$, divided by the total number of true positions of all query concepts. 

\item \textbf{Precision@}$k$ is the number of query concepts' true positions ranked in the top $k$, divided by the total number of queries times $k$.
 
\end{itemize}

\smallskip
\noindent \textbf{Compared Methods.}
To the best of our knowledge, we are the first to study taxonomy completion task and there is no directly comparable previous method. Thus, we adapt the following related methods to our problem setting and compare \TaxoExpan with them:
\begin{enumerate}[leftmargin=*, itemsep=0mm]
\item \textbf{Closest-Position}: A rule-based method which ranks candidate positions based on the cosine similarity:
\begin{displaymath}
\small
s(n_q, n_{p}, n_{c}) = \frac{\text{cosine}(\mathbf{x}_p, \mathbf{x}_q)+\text{cosine}(\mathbf{x}_c, \mathbf{x}_q)}{2}
\end{displaymath}

\item \textbf{Single Layer Model}: A model that scores tuple by a standard single layer neural network which inputs the concatenation of the concept embeddings. 
\item \textbf{Multiple Layer Model}: An extension of Single Layer Model that replaces the single layer neural network with multiple layer neural network.
\item \textbf{Bilinear Model} \cite{Sutskever09, Jenatton12}: It incorporates the interaction of two concept embeddings through a simple and efficient bilinear form. 
\item \textbf{Nerual Tensor Network} \cite{socher2013reasoning}: It incorporates Single Layer Model with a bilinear tensor layer that directly relates the two concept embeddings across multiple dimensions and a bias vector.
\item \textbf{TaxoExpan} \cite{taxoexpan}:
One state-of-the-art taxonomy expansion framework which leverages position-enhanced graph neural network to capture local information and InfoNCE loss\cite{Oord2018RepresentationLW} for robust training.
\item \textbf{ARBORIST} \cite{manzoor2020expanding}:
One state-of-the-art taxonomy expansion model which aims for taxonomies with heterogeneous edge semantics and optimizes a large-margin ranking loss with a dynamic margin function.

\end{enumerate}

Notably, except for the rule-based method Closest-Position, other baselines are learning-based method and designed for one-to-one matching. Thus we concatenate the embeddings of candidate's constituting concepts as candidate embedding to fit our one-to-pair setting. For fair comparison, we replace the GNN encoder of TaxoExpan with initial feature vector to align with other compared methods. There are other recently-proposed taxonomy expansion methods, \eg, HiExpan \cite{Shen2018HiExpanTT} and STEAM \cite{steam}. We do not include them as baselines because they leverage external sources, \eg, text corpus, to extract complicated features, while \TaxoExpan and other baselines only take initial feature vectors as input.

\smallskip
\noindent \textbf{Parameter Settings.}\label{subsubsec:implementation_details}
For learning-based methods, we use Adam optimizer with initial learning rate 0.001 and ReduceLROnPlateau scheduler\footnote{\scriptsize \url{https://pytorch.org/docs/stable/optim.html\#torch.optim.lr\_scheduler.ReduceLROnPlateau}} with ten patience epochs. During model training, the batch size and negative sample size is set to 128 and 31, respectively. We set $k$, \ie, the dimension of internal feature representation, to be 5. For \TaxoExpan, we simply set $\lambda_1=\lambda_2=\lambda_3=1$ to avoid heavy hyperparameter tuning.

\subsection{Experimental Results}  
\smallskip
\noindent \textbf{Overall Performance.}
Table~\ref{tbl:overall_results} presents the results of all compared methods on the four datasets. First, we find that learning-based methods clearly outperform rule-based Closest-Position method. 
Second, there is no baseline that could consistently outperform others in all taxonomies, which indicates the diversity of taxonomies of different domains and the difficulty of taxonomy completion task. 
Third, ARBORIST and TaxoExpan do not work well in taxonomy completion, which indicates that methods carefully designed for taxonomy expansion task will struggle in taxonomy completion task.
Finally, our proposed \TaxoExpan has the overall best performance across all the metrics and defeats the second best method by a large margin.

\begin{table}[h]
	\centering
	
	\scalebox{0.8}{
        \begin{tabular}{c|cccc}
        		\toprule
             	\multirow{2}{*}{\textbf{Method}} & \multicolumn{4}{c}{\textbf{MAG-Psychology}} \\
		\cmidrule{2-5}
		&MR &MRR & Recall@10  & Prec@1   \\
		\midrule
		TaxoExpan & 175.730  & 0.594  & 0.477   & 0.153 \\
		ARBORIST & 119.935  & 0.722   & 0.629  & 0.258 \\
 		\TaxoExpan   & 69.293 & 0.740 & 0.646  & 0.329 \\
		\bottomrule
         \end{tabular}
    }
	\scalebox{0.8}{
        \begin{tabular}{c|cccc}
        		\toprule
             	\multirow{2}{*}{\textbf{Method}} & \multicolumn{4}{c}{\textbf{WordNet-Verb}} \\
		\cmidrule{2-5}
		&MR&MRR & Recall@10  & Prec@1   \\
		\midrule
        TaxoExpan & 642.694  & 0.410 & 0.319   & 0.098  \\
        ARBORIST & 608.668  & 0.380  & 0.277 & 0.067 \\
 		\TaxoExpan    & 464.970 & 0.479  & 0.379  & 0.132\\
		\bottomrule
         \end{tabular}
    }
    \caption{Results on taxonomy expansion task.}
	\label{tbl:taxoexpan}
\end{table} 
\noindent \textbf{Performance on Taxonomy Expansion.}
As taxonomy expansion being a special case of our novel taxonomy completion task, we are curious about how \TaxoExpan performs on previous task. 
Thus, we compare \TaxoExpan with ARBORIST and TaxoExpan on taxonomy expansion task\footnote{\scriptsize We sample validation/test set from leaf nodes for taxonomy expansion task.}. The results are presented in Table \ref{tbl:taxoexpan}. Notice that ARBORIST and TaxoExpan is trained directly on taxonomy expansion task, while \TaxoExpan is trained solely on taxonomy completion task. 
From the results, we can see \TaxoExpan outperforms the others in both dataset with a large margin, which indicates that \TaxoExpan is able to solve taxonomy expansion task better than previous state-of-the-arts.

\begin{table}[h]
	\centering
	
	\scalebox{0.75}{
        \begin{tabular}{c|cccc}
        		\toprule
             	\multirow{2}{*}{\textbf{Method}} & \multicolumn{4}{c}{\textbf{MAG-Psychology}} \\
		\cmidrule{2-5}
		&MR&MRR & Recall@10  & Prec@10   \\
		\midrule
		\TaxoExpan w/o CG& 265.729 & 0.385   & 0.298  &  0.061 \\
		\TaxoExpan w/o $s_1$\&$s_2$ & 258.382   & 0.458 & 0.368 &  0.075 \\
		\TaxoExpan w/o $s_1$ & 269.058 & 0.471 & 0.382 & 0.123 \\
		\TaxoExpan w/o $s_2$ & 229.306 & 0.474 &  0.381  & 0.078 \\
		\TaxoExpan w/o $s_3$ & 342.021& 0.326 & 0.213   & 0.043\\
 		\TaxoExpan     &212.298 &	0.471  &0.377 &		0.077\\
		\bottomrule
         \end{tabular}
    }
	\scalebox{0.75}{
        \begin{tabular}{c|cccc}
        		\toprule
             	\multirow{2}{*}{\textbf{Method}} & \multicolumn{4}{c}{\textbf{WordNet-Verb}} \\
		\cmidrule{2-5}
		&MR&MRR & Recall@10  & Prec@10   \\
		\midrule

		\TaxoExpan w/o CG & 1578.120 	& 0.260&  0.178 		& 0.027\\
		\TaxoExpan w/o  $s_1$\&$s_2$    & 1497.843 & 0.278 	& 0.200 	& 0.030	\\
		\TaxoExpan w/o $s_1$ & 1530.192 & 0.293 & 0.219  & 0.033\\
		\TaxoExpan w/o $s_2$ & 1586.740 & 0.290 & 0.207 & 0.031 \\
		\TaxoExpan w/o $s_3$ & 1604.802  & 0.248 & 0.159  & 0.024\\
 		\TaxoExpan     &1445.801 	&0.304 &		0.215&		0.032\\
		\bottomrule
         \end{tabular}
    }
    \caption{Ablation analysis on MAG-Psychology and WordNet-Verb datasets.}
	\label{tbl:ablation}
\end{table} 
\noindent \textbf{Ablation Study.}
We conduct the ablation studies on two representative datasets \textbf{MAG-Psychology} and \textbf{WordNet-Verb}, and the results are presented in Table \ref{tbl:ablation}. The results show that without any of the key components of \TaxoExpan, \ie, auxiliary scorers ($s_1$, $s_2$ and $s_3$) and channel-wise gating mechanism (CG), the overall performance will degrade by different extends, which indicates the effectiveness of the components.

\begin{table}[h]
	\centering
	\scalebox{0.7}{
        \begin{tabular}{c|c|c}
        		\toprule
             	\textbf{Dataset} &\textbf{\# of candidate pairs} &\textbf{Avg. running time per query (s)}\\
		\midrule
		\textbf{MAG-CS} &153,726&0.131\\
		\textbf{MAG-Psychology} &101,077& 0.067\\
		\textbf{WordNet-Verb} & 51,159 & 0.055\\
		\textbf{WordNet-Noun} & 799,735 & 0.870\\
		\bottomrule
         \end{tabular}
    }
    \caption{Efficiency Analysis.}
	\label{tbl:eff}
\end{table} 
\smallskip
\noindent \textbf{Efficiency Analysis.}
At the training stage, our model uses $|\N^{(0)}|$ training instances every epoch and thus scales linearly to the number of concepts in the existing taxonomy. At inference stage, because the cardinality of candidate pairs is $|\N^{(0)}|^2$ without any restriction, for each query concept, we need to calculate $|\N^{(0)}|^2$ matching scores, one for every candidate pair. However, in practical, as we restrict the valid candidate pairs to be $\langle$ancestor, descendant$\rangle$ concept pairs in existing taxonomy, the number of candidates need to be considered is substantially reduced and therefore the inference efficiency is largely improved. Also, the inference stage can be further accelerated using GPU. We list the number of valid candidate pairs and the average running time per query during inference stage of all datasets in Table \ref{tbl:eff}. From the table, we can see the number of valid candidate pairs is no more than ten times of $|\N^{(0)}|$ and thus the inference stage is quite efficient. 

\begin{figure}[h]
  \centering
  \centerline{\includegraphics[width=0.95\linewidth]{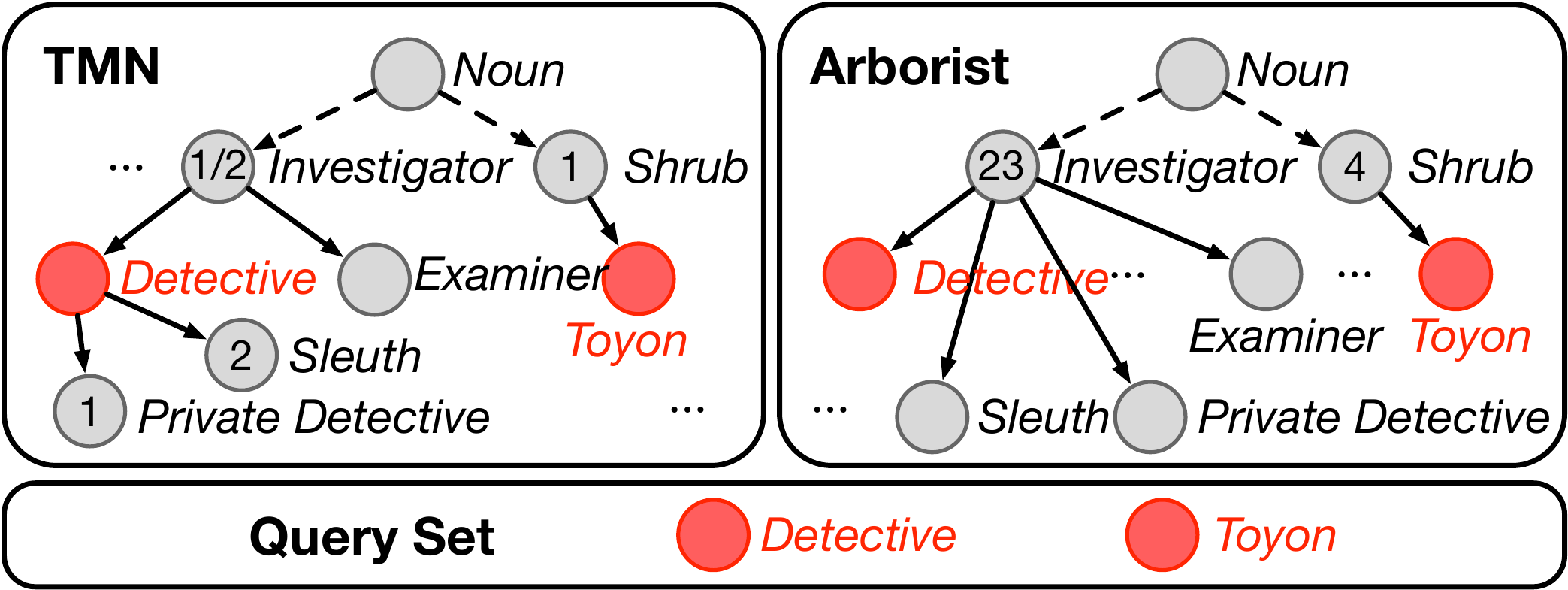}}
  \caption{Case Study. The grey concepts are existing concepts while the red ones are queries needed to be inserted. The dash lines mean an omitting of some internal nodes and the solid lines indicate real edges in taxonomy. The numbers inside nodes are the ranking position output by the model. We can see \TaxoExpan recovers the true positions for both internal and leaf concepts.}
  \label{fig:case study}
\end{figure}

\noindent \textbf{Case Study.}
We illustrate the power of \TaxoExpan via two real query concepts \mquote{Detective} and \mquote{Toyon} of \textbf{WordNet-Noun} in Fig.\ref{fig:case study}. For internal concept \mquote{Detective}, \TaxoExpan ranks the true positions $\langle$\mquote{Investigator}, \mquote{Private Detective}$\rangle$ at top 1 and $\langle$\mquote{Investigator}, \mquote{Sleuth}$\rangle$ at top 2, while Arborist can only rank the true parent \mquote{Investigator} at top 23. For leaf concept \mquote{Toyon}, \TaxoExpan recovers its true parent \mquote{Shrub} but Arborist ranks \mquote{Shrub} at top 5. We can see that \TaxoExpan works better than baseline in terms of recovering true positions. 
\section{Related Work}\label{sec:related_work}
\noindent \textbf{Taxonomy Construction and Expansion.} Automatic taxonomy construction is a long-standing task in the literature. Existing taxonomy construction methods leverage lexical features from the resource corpus such as lexical-patterns \cite{Nakashole2012PATTYAT,Jiang2017MetaPADMP, Hearst1992AutomaticAO,Agichtein2000SnowballER} or distributional representations \cite{P18-1229, Zhang2018TaxoGenCT, Jin2018JunctionTV,Luu2016LearningTE,Roller2014InclusiveYS,Weeds2004CharacterisingMO} to construct a taxonomy from scratch. However, in many real-world applications, some existing taxonomies may have already been laboriously curated and are deployed in online systems, which calls for solutions to the taxonomy expansion problem. To this end, multitudinous methods have been proposed recently to solve the taxonomy expansion problem~\cite{Vedula2018EnrichingTW, Shen2018HiExpanTT, manzoor2020expanding, taxoexpan, steam,mao2020octet}. For example, 
Arborist \cite{manzoor2020expanding} studies expanding taxonomies by jointly learning latent representations for edge semantics and taxonomy concepts;
TaxoExpan \cite{taxoexpan} proposes position-enhanced graph neural networks to encode the relative position of terms and a robust InfoNCE loss;
STEAM \cite{steam} re-formulates the taxonomy expansion task as a mini-path-based prediction task and proposes to solve it through a multi-view co-training objective. However, all the existing taxonomy expansion methods aim for solving the one-to-one matching problem, \ie to find the true parent/hypernym, which is incompatible to our novel one-to-pair matching problem induced by taxonomy completion task.

\noindent \textbf{Auxiliary Learning.} Auxiliary learning refers to a learning strategy that facilitates training of a primal task with auxiliary tasks \cite{ruder2017overview, shen2018multi}. Different from multi-task learning, auxiliary learning only cares the performance of the primal task. The benefits of auxiliary learning have been proved in various applications \cite{which_tasks, tang2020improving, trinh2018learning, toshniwal2017multitask, hwang2020selfsupervised, jaderberg2016reinforcement, pmlrv70odena17a, liu2019self, lin2019adaptive, xiao2019similarity}. In most of these contexts, joint training with auxiliary tasks adds an inductive bias, encouraging the model to learn meaningful representations and avoid overfitting spurious correlations. 
Despite the numerous applications of auxiliary learning, its benefits on taxonomy construction remains less investigated. To our best knowledge, we are the first to leverage auxiliary learning to enhance taxonomy construction.

\section{Conclusion}\label{sec:conclusion}
This paper studies taxonomy completion without manually labeled supervised data. 
We propose a novel \TaxoExpan framework to solve the one-to-pair matching problem in taxonomy completion, which can be applied on other applications where one-to-pair matching problem exists.  
Extensive experiments demonstrate the effectiveness of \TaxoExpan on various taxonomies. 
Interesting future work includes leveraging current method to cleaning the existing taxonomy, and incorporating feedback from downstream applications (\eg, searching \& recommendation) to generate more diverse (auxiliary) supervision signals for taxonomy completion.

\section{Acknowledgements}\label{sec:ack}
Thanks the anonymous reviewers for their helpful comments and suggestions. We also thank Jingjing Xu, Hao Zhou, and Jiawei Han for their valuable discussions.
\bibliography{cite}

\end{document}